# Urdu Poetry Generated by Using Deep Learning Techniques

**Muhammad Shoaib Farooq, Ali Abbas**
Department of Computer Science, School of System and Technology, University of Management and Technology, Lahore, 54000
Corresponding author: Muhammad Shoaib Farooq (Shoaib.farooq@umt.edu.pk)

## Abstract
This study provides Urdu poetry generated using different deep-learning techniques and algorithms. The data was collected through the Rekhta website, containing 1341 text files with several couplets. The data on poetry was not from any specific genre or poet. Instead, it was a collection of mixed Urdu poems and Ghazals. Different deep learning techniques, such as the model applied Long Short-term Memory Networks (LSTM) and Gated Recurrent Unit (GRU), have been used. Natural Language Processing (NLP) may be used in machine learning to understand, analyze, and generate a language humans may use and understand. Much work has been done on generating poetry for different languages using different techniques. The collection and use of data were also different for different researchers. The primary purpose of this project is to provide a model that generates Urdu poems by using data completely, not by sampling data. Also, this may generate poems in pure Urdu, not Roman Urdu, as in the base paper. The results have shown good accuracy in the poems generated by the model.
*Keywords*: Urdu Poetry, Poetry Generation, LSTM, GRU, Deep Learning.

## Introduction

Artificial Intelligence (AI) has helped computers change significantly over the past few decades, from simple tools to smart machines. The fourth industrial revolution witnessed the pervasive application of artificial intelligence across numerous industries, with observable effects on self-driving vehicles, business analysis, self-learning machines, medical report analysis, market exposure, and complex problems. Deep learning NLP techniques have enabled AI to analyze human language and recognize images. Recent years have seen a notable increase in research into language processing, particularly in Urdu, including sentiment analysis, writing styles, and poetry generation. However, the field of Urdu poetry creation still requires investigation. The purpose of this paper is to produce Urdu poetry in Sales Urdu. The paper falls under the generic AI domain and the sub-domain of Natural Language Processing. The literature review reveals a substantial gap in the generation of Urdu text, which is the primary objective of this paper. Urdu's distinctive writing style is another factor, as it is essential for the model's training. The other essential but challenging aspect is data acquisition and cleansing. The methodology section will provide discussion on different methods along with their technical definitions and equations that are going to be used in this model.

## Literature review

Applying deep learning techniques to generate poetry in various languages has gained significant attention recently. This review explores the landscape of Urdu poetry generation using deep learning, focusing on the methodologies employed and the uniqueness of the dataset used. The literature review analyzes the work of other researchers in computational creativity, explicitly focusing on poetry generation using deep learning techniques. The selected literature spans the last 10 years and is sourced from reputable platforms such as JSTOR, IEEE Xplore, ScienceDirect, and SpringerLink. Various papers have discussed implementing BiLSTM, GPT, GPT-2, LDA, LSI, VSM, RNN, and CNN for poetry generation using different datasets. Additionally, the review highlights research on multilingual poetry generation, encompassing Urdu, Arabic, English, Roman, and Hindi. The base paper's primary focus is on deep learning-based Urdu poetry generation while acknowledging the limitation of limited work on pure Urdu poetry instead of Roman Urdu.

A paper by [1] analyzes machines' historical creativity and the advancements that have improved their computational methods. The paper focuses on the enhanced computational creativity of machines, exemplified by machine learning techniques. These machines can now generate and evaluate new artifacts based on given data. The study covers various levels of creativity, including autonomous evaluation, change, and non-randomness. It explores learning to evaluate using regression or classification techniques and the algorithms used in these models.

[2] The paper delves into machine creativity, particularly in generating poetry through machine learning algorithms. They sourced data from online repositories, such as Gutenberg, containing over 50 thousand eBooks. Emotion poems were classified using the GPT-2 architecture, evoking emotions like sadness, anger, anticipation, joy, trust, and dreams. Five showed high scores after training eight emotion words and comparing them with subsets of the corpus. The top 20 poems, out of 1000, were chosen based on GPT scores and reviewed by ten native English speakers.

[3] researchers have recently turned their focus to emotion classification from web information. While much of the existing research has centered on categorizing emotions from informal text like chat, text messages, and social media content, less attention has been paid to emotion classification from formal literature, such as poetry. The paper introduces a deep neural network model, specifically a BiLSTM model, for emotion classification from poetry text. They evaluate the model's performance using a benchmark poetry dataset and successfully categorize poems into emotional types: love, anger, loneliness, suicide, and surprise. The proposed model's effectiveness is compared against various baseline methods, including machine learning and deep learning models. [4] utilizes neural memory to generate creative Chinese poetry, acknowledging that statistical methods alone are insufficient for poetry generation. The paper incorporates machine learning, employing memory augmentation alongside RNN implementation. [5] Deep learning was applied for sentiment analysis using poetry from various online sources. The author employed a bidirectional Long Short-Term Memory (LSTM) model with CNN. The model achieved a peak accuracy of 88%. [6] The paper uses machine learning to classify Arabic poetry based on different emotions. They have Arabic poetry data from online resources and implemented SVM, Naive Bayes, voting feature intervals, and hyper pipes.

[7] A paper worked on a neural network-based technique for producing Urdu and Hindi poetry. The model has been trained on an Urdu and Hindi poetry corpus, enabling it to generate compositions with the correct meter and rhyme scheme. [8] Another paper presents a model for creating poetry using deep learning techniques and enhanced phonetic and semantic embeddings.

Similar study done by [9] which analyzes the data collection of Urdu poetry. This analysis concentrates on the distribution of words, meters, and rhythms throughout the corpus. This paper [10] presents a technique for writing Urdu poetry using RNNs and LSTMs models. The model has been instructed in an Urdu poetry collection. In addition, the paper examines the challenges of writing poetry in Indic languages, such as their complex morphology and syntax. [11] The model has been trained in a compilation of Urdu poetry and can write poems with the correct meter and rhyme scheme.

[12] The study examines the significance of social media as a platform for sharing content and the difficulty of comprehending how individuals communicate and express their opinions about particular products. The study focuses on Urdu sentiment analysis due to the language's widespread use, presenting the system architecture and experimental approaches and achieving a verified accuracy of 66 percent in sentiment analysis of Urdu comments on multiple websites.

[13] Presents a research study that utilizes a text classification algorithm to identify poetry styles. Knowledge representation algorithms are employed for text classification, enhancing the intelligent analysis of poetry styles. The study combines various intelligent algorithms to develop an Urdu poetry style analysis system, including preprocessing the poetry documents in the corpus and mapping them into vector forms accessible by a computer. Simulation tests are conducted to verify the effectiveness of the poetry style analysis system based on the text classification method.

[14] Explores the use of natural language processing and computational creativity to generate poetry in the Ottoman Turkish language through machine learning.

The data was collected from OTAP, and a vocabulary was created. The paper addresses three main limitations. To address these challenges, the author used a Finite State Transducer (FST) before training the RNN recurrent neural network model, aiming to connect different syllables in the poem. However, this approach has some limitations in generating variations with a single word, for which the FSA model may be used to score the words.

[15] The paper focuses on interactive poetry generation using a model similar to the previous study. The author integrates recurrent neural networks with the Finite State Acceptor. They discuss the development of an app for real-time poetry generation using machine learning models, where users input a topic, and the app generates a poem accordingly. The real-time nature of the process poses challenges, including slower generation capability due to on-the-spot topic input, limited evaluation of poem quality, and the model's inability to adjust the generated poem. However, the app offers configurable tools to edit poems, providing an interactive experience.

The research was done by [16] on the generation of Urdu Poetry using the NLP technique for research in neurocognitive poetry and styling of poetry on computers. Data for this study is collected from Rekhta, a famous Urdu poetry platform. The study uses the data of 48,761 words of ghazals produced by 4,754 poets in 800 years. The study uses Multidimensional Scaling to numerically analyze the similarities and differences between different poetic works done by different well-known poets.

The authors of this paper [17] have analyzed the emotions in the Roman Urdu text from the data of approximately 10,000 sentences. The authors have created a corpus of data. They have used a rule-based, recurrent convolutional neural network model and N-gram to evaluate the sentiment analysis on the corpus dataset. The results from the RCNN model have been compared using Rule-based and N-gram models.

[18] have also discussed similar models and generated English and Romanian poetry through machine learning experiments. The author has used a quantitative approach and provided a statistical experiment to compare two poetical texts using information entropy and N-gram informational energy. After this statistical analysis, the author jumped into a deep neural network model that RNN has used to generate Byron's poetry. Another paper has used Markov chains and LSTM RNN models to generate poetry in the Romanian language. The authors [19] have focused on generating poetry rather than based on emotions, as all did in the previous papers discussed above.

This paper by [20] has worked on classifying Urdu Ghazals using 3000 different Ghazals datasets. Urdu poetry has a vast and complex poetry structure because it has a connection back to Arabs. In this paper, after the data preprocessing, the authors have applied different machine learning techniques such as support vector machine, random forest, decision tree, Naive Bayes, and KNN models to identify the Urdu ghazals. Similar work has been done by [21] using the same machine-learning models. Different genres of poems were taken in the Arabic language. The author has discussed that preprocessing was an essential step for this machine learning model as it has increased classification accuracy. The highest average accuracy obtained from the three models was 51 % from the linear support vector classification. Here is another paper by [22], which uses 450 Hindi poem data to do classification using supervised machine learning using R language. They have used random forest and naive Bayes models to compare the misclassification error in the corpus dataset. They have 83% accuracy in their model.

The research done by [23] uses a Handwritten Urdu Character Dataset; this dataset is found from the Nastaliq Urdu script. The information was acquired from 750 Kashmir Valley residents and included solitary and positioned characters and digits.

These authors [24] have presented a paper in which they have performed different unsupervised machine learning models on different datasets to compare their accuracy in results. They have used three datasets collected from news articles: Allam Iqbal's and other poets' poetry. They have done tokenization, lemmatization, removing stop words, and stemming on the corpus dataset.

[25] His paper has combined statistical analysis with the NLP to learn about the architecture of poetry generation. This paper uses the following statistical models: LSI, LDA, and VSM. This has shown promising results from the lantern pattern recognition. Research done by [26] shows the extraction of semantic and logical topics from any text collection; machine learning-based approaches proved helpful. Different topic modeling approaches are used for Urdu

poetry to show that these approaches are equally helpful for text generation.

These are just a few of the papers that have been published on Urdu poetry generation using ML models. The field is still in its early stages, but there has been significant progress in recent years. As the models become more sophisticated, they can generate more creative and semantically meaningful poetry. Overall, the literature review highlights the progress and potential of using NLP and deep learning techniques for creative applications in various languages, providing valuable insights into computational creativity.

# Statement of the problem and significance of the study

## Problem statement

This paper will focus on the Urdu Poetry generation with the help of Deep Neural Networks and Natural Language processing models such as N-Gram, Bi-Gram models, Gated Recurrent Units, and Long Short-term Memory Networks.

## Scope of the study

**Scope:**

- The study focuses on generating poetry in Urdu Salees, distinct from Roman Urdu, a relatively less explored area in poetry generation research.
- The research encompasses multiple genres of poetry, creating a diverse dataset for analysis and generation.

Limitations:

- The data used in the study is anonymous and not associated with any specific poet, which might limit the ability to analyze poetry from individual authors.
- The study utilizes the complete dataset, forgoing sampling methods, which might impact the generalization of results.

## Significance of the study:

The current research uses creative computing methods to generate poetry in different languages. NLP models have significantly simplified the analysis of business customer comments and reviews. Data Science has accelerated data analysis and pattern recognition, making it more efficient than traditional methods. Intelligent machines are now being taught languages and generating new content, including poetry. The paper aims to generate Urdu poetry based on renowned Urdu poets' styles and contribute to the field of Artificial Intelligence. The objective is to replicate the writing styles of legendary poets with high precision and enable computers to create poetry for everyone. This research will revolutionize Urdu poetry and make it easier to write poetry in the manner of renowned authors such as Mirza Ghalib.

## Methodology

This study employs a Recurrent Neural Network (RNN), a standard machine learning model for language and sequence analysis, in its investigation. Before delving into RNN, the Forward Feed Neural Network is discussed, which consists of unprocessed input, hidden, and output layers. However, this model's requirement to consider previous inputs renders it inappropriate for detecting sequences in language processing. RNN overcomes this limitation by preprocessing and storing the previous input state, enabling it to evaluate both previous and current language states when predicting the next word following a given word.

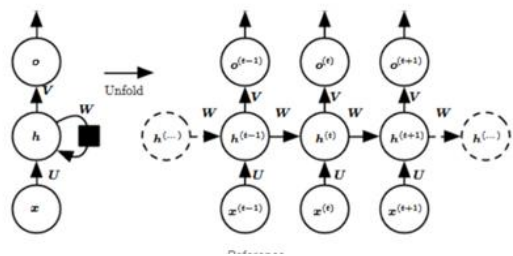

The RNN model has an initial concealed state (h0) and receives input at every time step. The input and the previous hidden state (h(t-1)) are processed by the hidden layer to produce the current hidden state (h(t)). The current concealed state is output and passed as the prior hidden state to the following time phase. This procedure is repeated for each input at various time intervals. The model employs shared weights (U, V, W) to process all concealed layers efficiently.

**Forward Pass**

We presume a hyperbolic tangent activation function for the hidden layer in the illustration. Since RNN is used to predict words and characters, we regard the output as unnormalized log probabilities for each possible value of the discrete variable. As a post-processing phase, the SoftMax operation can express discrete variables as a vector of normalized probabilities over the output.

RNN forward pass is represented by the below equation:

$$a^{(t)} = b + Wh^{(t-1)} + Ux^{(t)}$$
$$h^{(t)} = \tanh(a^{(t)})$$
$$o^{(t)} = c + Vh^{(t)}$$
$$\hat{y}^{(t)} = \text{softmax}(o^{(t)})$$

The input sequence's duration corresponds to the output signal's length in this example. The entire loss of all x values coupled with a set of y values equals the sum of all losses across all time increments. The Softmax function acquires the vector of probabilities over the output by receiving the outputs as an operand. Given the current input, the loss L is defined as the negative log probability of the correct target.

**Backward Pass**

The forward propagation pass traverses the graph from left to right, while the backward propagation pass computes the gradient from right to left. Because the forward propagation graph is sequential, parallelization cannot reduce the runtime, which remains O(τ). The memory cost is also O(τ) because the states computed during the forward pass must be stored until used during the rearward run. Backpropagation through time is accomplished by applying backpropagation to an unrolled graph at a cost of O(τ).

**Computing Gradients**

We have been provided with a loss function denoted by L; we must compute the gradients for our three-weight metrics, U, V, W, and the bias terms B and C. Then, we update them with an L-based learning rate. Like conventional backpropagation, the gradient gives us an idea of how the loss varies for each weight parameter.

We revise the weights W using the following equation to minimize loss:

$$W \leftarrow W - \alpha \frac{\partial L}{\partial W}$$

Nodes in our computational graph (t) include the parameters U, V, W, b, and c, as well as the sequence of nodes indexed by t for x (t), h(t), o(t), and L. We must recursively calculate the gradient nL for each node n in the graph based on the gradient obtained at nodes that follow it. The output gradient is determined using o(t) as the SoftMax function argument to generate the vector y of probabilities over the output.

$$(\nabla_{o^{(t)}} L)_i = \frac{\partial L}{\partial o_i^{(t)}} = \frac{\partial L}{\partial L^{(t)}} \frac{\partial L^{(t)}}{\partial o_i^{(t)}} = \hat{y}_i^{(t)} - 1_{i=y^{(t)}}$$

As shown in the diagram below, the hidden state h(t) has a gradient that flows from both the present output and the future concealed state at time t.

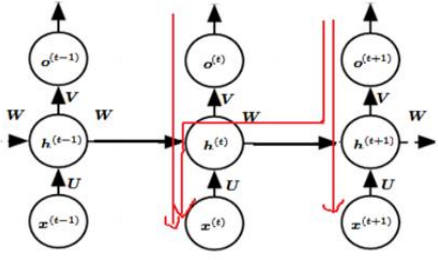

We commence after the sequence and proceed backward. At the final time step, h(τ) has only O(τ) as a descendant; therefore, its gradient is straightforward.

$$\nabla_{h^{(\tau)}} L = V^\top \nabla_{o^{(\tau)}} L$$

We can then repeat backward in time, from t=1 to t=1, to back-propagate gradients across time, noticing that h(t) for t has both o(t) and h(t+1) as descendants.

The gradient is given below.

$$\nabla_{h^{(t)}} L = \left(\frac{\partial h^{(t+1)}}{\partial h^{(t)}}\right)^\top (\nabla_{h^{(t+1)}} L) + \left(\frac{\partial o^{(t)}}{\partial h^{(t)}}\right)^\top (\nabla_{o^{(t)}} L)$$
$$= W^\top \text{diag}\left(1 - \left(h^{(t+1)}\right)^2\right)(\nabla_{h^{(t+1)}} L) + V^\top (\nabla_{o^{(t)}} L)$$

After obtaining the gradients on the interior nodes of the computational graph, we can obtain the gradients on the parameter nodes.

$$\nabla_c L = \sum_t \left(\frac{\partial o^{(t)}}{\partial c}\right)^\top \nabla_{o^{(t)}} L = \sum_t \nabla_{o^{(t)}} L$$
$$\nabla_b L = \sum_t \left(\frac{\partial h^{(t)}}{\partial b^{(t)}}\right)^\top \nabla_{h^{(t)}} L = \sum_t \text{diag}\left(1-\left(h^{(t)}\right)^2\right)\nabla_{h^{(t)}} L$$
$$\nabla_V L = \sum_t \sum_i \left(\frac{\partial L}{\partial o_i^{(t)}}\right)\nabla_{V^{(t)}} o_i^{(t)} = \sum_t (\nabla_{o^{(t)}} L) h^{(t)\top}$$
$$\nabla_W L = \sum_t \sum_i \left(\frac{\partial L}{\partial h_i^{(t)}}\right)\nabla_{W^{(t)}} h_i^{(t)} = \sum_t \text{diag}\left(1-\left(h^{(t)}\right)^2\right)(\nabla_{h^{(t)}} L) h^{(t-1)\top}$$
$$\nabla_U L = \sum_t \sum_i \left(\frac{\partial L}{\partial h_i^{(t)}}\right)\nabla_{U^{(t)}} h_i^{(t)} = \sum_t \text{diag}\left(1-\left(h^{(t)}\right)^2\right)(\nabla_{h^{(t)}} L) x^{(t)\top}$$

Using the chain rule, the gradient calculations for each parameter are as follows:

### Issues

The RNN is a strong and straightforward model in theory but challenging to train effectively. The vanishing gradient and exploding gradient issues are vital reasons for its difficulties. Backpropagation Through Time (BPTT) training requires gradient propagation from the terminal to the initial cell. The sum of these gradients can either become zero or develop exponentially, resulting in the problem of expanding gradients, in which the gradient norm substantially increases during training.

### LSTM Model

Our research uses the RNN subtype, LSTM (Long et al.), as we are working on Urdu poetry generation. We need a model that can process the sequence of the cell, so LSTM is a perfect fit for our model. It detects which words in Urdu poetry will come next to maintain the beauty of poetry.

The LSTM model consists of three gates. The first component is the Forget Gate, followed by the Input Gate and the Output Gate. In the neglect gate, information is processed, such as which information in a sentence must be essential and which is unnecessary. The irrelevant information is wiped out, and the excellent and essential information is passed to the input layer. Then, in the input, the data is processed for the following upcoming output. Therefore, this project is beneficial in language processing tasks.

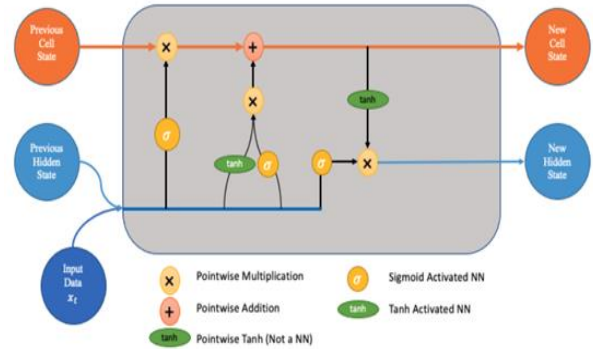

An LSTM network can learn this pattern, which occurs every 12 time periods. It does not merely utilize the previous forecast; it also maintains a longer-term context, which helps it avoid the issue of other models' long-term reliance.

## Working Analysis

As mentioned, LSTM employs a series of gates to process information effectively. A typical LSTM comprises three gates: neglect, input, and output.

## Forget Gate

The disregard gate is the beginning of the procedure. We will determine which portions of the cell state are advantageous using the previous concealed state and new input data. Consider each element of this vector as a filter that allows more data to pass as its value approaches 1. The transferred values are then multiplied point-by-point with the previous cell state.

## Input Gate

This stage attempts to determine what new data should be recorded in the network's long-term memory based on the previously concealed state and new input data. The new memory network and the input gate are neural networks that receive the same inputs, the previous concealed state, and the new input data. The new memory network is an activated neural network that has learned how to build a new memory update vector by combining the existing concealed state with new input data. Given the prior concealed state context, this vector comprises information from the new input data. Given the new data, this vector indicates how much each component of the network's long-term memory should be updated. The input gate is a sigmoid-activated network that determines which portions of the new memory vector will be retained.

## Output Gate

The output gate is responsible for determining the new concealed state. This decision will be based on the recently updated cell state, the prior concealed state, and the new input data. One might assume that we could output the updated cell state; however, this would be akin to someone sharing everything they know about the stock market when asked whether it will rise or fall tomorrow.

## GRU Model

GRUs are a superior form of recurrent neural networks. GRU utilizes the update and reset gates to circumvent the RNN's vanishing gradient problem. Two vectors determine the data to be transmitted to the output. They are exceptional because they can be taught to retain knowledge from the past without losing it over time or eliminating irrelevant data.

## Update Gate

The update gate helps the model determine how much information from previous time steps should be carried forward to subsequent time steps. This is extremely useful because the model can replicate all historical data, eradicating the possibility of gradients disappearing. We will later examine how to utilize the update gate.

## Reset Gate

The model uses this gate to determine how much historical knowledge should be forgotten.

This is identical to the formula for the update gate. The distinction lies in the weights and usage of the gate, which will be discussed subsequently. The restored gate is depicted in the following diagram.

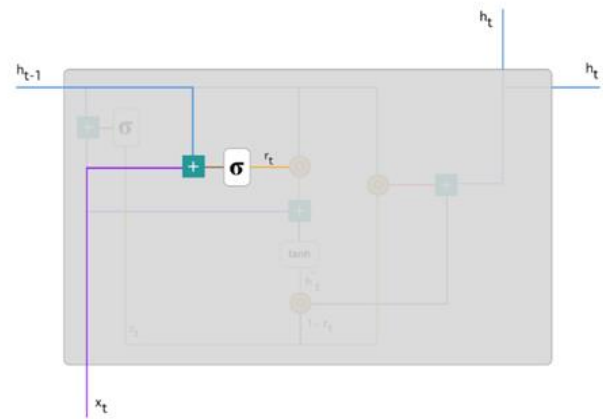

## Working

The initial step is to establish a candidate concealed state.

$$\hat{H}_t = \tanh(x_t * U_g + (r_t \circ H_{t-1}) * W_g)$$

It multiplies the input and hidden state from the previous timestamp, t-1, by the output of the reset gate rt. The concealed state of the candidate is determined by passing all of this information to the tanh function.

$$\hat{H}_t = \tanh(x_t * U_g + (r_t \circ H_{t-1}) * W_g)$$

The most important aspect of this equation is how we use the value of the reset gate to determine how much the preceding concealed state can influence the candidate state.

## Hidden State

The candidate state, once obtained, generates the concealed state Ht. Here, the Update checkpoint enters into action. Unlike LSTM, GRU utilizes a single update gate to govern historical information (Ht-1) and the most recent information.

$$H_t = u_t \circ H_{t-1} + (1-u_t) \circ \hat{H}_t$$

If it is close to zero, the first term in the equation will vanish, indicating that the new hidden state will not contain much information from the previous one. The second portion merges into one, indicating that the current concealed state will only contain information from the candidate state.

$$H_t = u_t \circ H_{t-1} + (1-u_t) \circ \hat{H}_t$$

Similarly, if the value of it is on the second term, the current concealed state will depend solely on the first term.

## GRU

GRU is also known as a gated neural network. It solves vanishing gradient problems that arise in standard recurrent neural networks. GRU is an improved version of LSTM because both are constructed identically and, in some instances, produce identical results. GRU is a more sophisticated variant of Recurrent Neural Network. It employs a reset gate and an update gate. These two vectors determine which data is transmitted to the output. They are helpful because they can be programmed to store data for an extended period, eliminating the need to delete unnecessary data.

## Update Gate:

The update gate helps the model determine how much historical data from previous time stages should be transmitted to the future. This is especially effective because the model can replicate all historical data and eliminate the possibility of gradient issues with vanishing values.

## Reset Gate:

The model utilizes this gate to determine how much historical data should be forgotten. We insert the h (t-1) and x t lines, multiply them by their respective weights, add the outcomes, and then flatten the curves with the sigmoid function.

## Final Memory in the Present State

If, during the sentiment analysis of a movie review, the most important information is contained solely in the first line and all other information is irrelevant, then our model can extract the sentiment from the first line and disregard the remaining text. It accepts the current state's input, which triggers the Update gate. Therefore, the Update gate is required to complete this concluding stage.

## Difference between GRU and LSTM:

Long short-term memory refers to the RNN architecture utilized in deep learning. LSTM networks are ideally adapted for processing, categorizing, and forming forecasts based on time string data, as there may be indeterminate time intervals between essential events in a time string. In 2014, a solution was developed for the typical recurrent neural network issue of declining gradient. GRU and LSTM share many similarities. Both algorithms regulate the process of memorization using a gating technique.

Interestingly, GRU is less complicated and computes much faster than LSTM. The fundamental difference between GRU and LSTM is that GRU has three gates—input, output, and forget—whereas LSTM has only two gates—reset and update. GRU has fewer gates than LSTM, which makes it simpler. GRU is superior to LSTM since it can be modified easily and does not require memory units. As a result, it is quicker to train than LSTM and give according to performance. GRU is used when the data series is small; LSTM is used when the data series is more extensive. GRU reveals the veiled layers and the whole memory but not by LSTM. LSTM and GRU are required for complex problem domains such as machine translation, speech recognition, speech synthesis, sentiment analysis, stock price prediction, and machine comprehension.

## Results and Conclusion

This section presents the results of various deep learning models applied to the Urdu poetry generation problem statement. The models utilized include LSTM and GRU. Specifically, we focus on the outcomes achieved with the GRU model across different epochs.

میں مرا سودا

صحرا اندازۂ ترا شاید جنوں اے تھا غلط
ہیں سکتے توڑ کو ہو و رنگ طلسم اس سے خودی
ویں شہر جانے نہ ہے کا طرح غم تعلقِ یہی
نہ سمجھتی پہ ساز ایک اک ہیں آئے وہ

*Output*:
ملتا نہیں کامل یا نقص کچھ گیا
ہے اتنا فرق میں نئی اور میں روشنی پرانی
نہیں انداز کا ساحل اسے ملتی نہیں کشتی اسے
ہوں میں کہ ہے کیا نے میں کے نام کی اس کہ ہے کیا
دی ہے آنکھیں مری تو کیا کیا مجھے یہ

*Output*:
ہو کیوں آسماں کا اس دشمن کے جس
ہیں کہتے کو کس ستانا تو آزمانا ہے یہی
ہے کیا کیا میرا تو تم جب لیے ہو کے عدو
کی نام کی اس بھی کو بازی مرے ہیں میں
میرا ہے کیا کیا تو اب ہے کی اس
نہیں میری نے میں

*Output*:
ہوئے مارے
نہیں میں مدفن اپنے گرد سوا کے گل جلوۂ
کا ناسور نئے ہے بپولیٰ اک قطرہ قطرہ
کی بات کی ان کر دیکھ ذوق بھی خوں
کی بات کی بات کی بات کی اس
کی بات کی بات کی بات کی اس
با کی بات کی بات کی اس

*Output*:
ہیں گئے جھلس منظر سارے کہ میں دوپہر ہے برسی آگ وہ
ہو نہ کہیں اب تازگی وہ تھی تازگی جو سویرے یہاں
بات کی بازی مرے ہے کیا کیا تو بھی کے اس
کی بات کی دور کی بات کی بات کی اس
کیا تو اب ہے کی اس

*Output*:
گزری کیا پہ ان بعد مرے کہ نہیں پتا
تھا آیا چھوڑ میں زمانے خواب چند میں
ہے کیا کیا تو بھی سے کوہ سہی ہی ذرے

ہے کیا کیا میرا ہے کیا کیا نے میں

کیا کیا نے میں

The results of the model provide significant results. Further modifications can be made as the era of generative AI is started, further approaches can be used, and results can be refined further.